\documentclass[preprint,12pt]{elsarticle}

\usepackage{amssymb}

\usepackage{amsmath}

\journal{Energy}

\usepackage{enumitem}
\usepackage[colorlinks=true, linkcolor=blue, citecolor=blue, urlcolor=blue]{hyperref}
\usepackage{algorithm}
\floatname{algorithm}{Algorithm}
\usepackage{algorithmic}

\usepackage{url}
\biboptions{sort&compress}
\usepackage{cleveref}
\crefname{algorithm}{Algorithm}{Algorithms}
\Crefname{algorithm}{Algorithm}{Algorithms}
\usepackage{multirow}
\usepackage{booktabs}
\usepackage{graphicx}
\usepackage{caption}
\usepackage{adjustbox}
\usepackage{array}
\crefname{figure}{Fig.}{Figs.} 
\Crefname{figure}{Fig.}{Figs.}

\begin{document}

\begin{frontmatter}

\title{Integrated Forecasting of Marine Renewable Power: An Adaptively Bayesian-Optimized MVMD-LSTM Framework for Wind-Solar-Wave Energy}

\author[label1,label2]{Baoyi Xie}
\author[label1,label2]{Shuiling Shi\corref{cor1}}
\author[label1,label2]{Wenqi Liu\corref{cor1}}

\affiliation[label1]{organization={Data Science Research Center},
            addressline={Kunming University of Science and Technology}, 
            city={Kunming},
            postcode={650500}, 
            country={China}}
            
\affiliation[label2]{organization={Faculty of Science},
	addressline={Kunming University of Science and Technology}, 
	city={Kunming},
	postcode={650500}, 
	country={China}}

\cortext[cor1]{Corresponding author: 20202111110@stu.kust.edu.cn (S. Shi), 11305004@kust.edu.cn (W. Liu).}

\begin{abstract}

Integrated wind–solar–wave marine energy systems hold broad promise for supplying clean electricity in offshore and coastal regions. By leveraging the spatiotemporal complementarity of multiple resources, such systems can effectively mitigate the intermittency and volatility of single-source outputs, thereby substantially improving overall power-generation efficiency and resource utilization. Accurate ultra-short-term forecasting is crucial for ensuring secure operation and optimizing proactive dispatch. However, most existing forecasting methods construct separate models for each energy source, insufficiently account for the complex couplings among multiple energies, struggle to capture the system’s nonlinear and nonstationary dynamics, and typically depend on extensive manual parameter tuning—limitations that constrain both predictive performance and practicality. We address this issue using a Bayesian-optimized Multivariate Variational Mode Decomposition–Long Short-Term Memory (MVMD–LSTM) framework. The framework first applies MVMD to jointly decompose wind, solar and wave power series so as to preserve cross-source couplings; it uses Bayesian optimization to automatically search the number of modes and the penalty parameter in the MVMD process to obtain intrinsic mode functions (IMFs); finally, an LSTM models the resulting IMFs to achieve ultra-short-term power forecasting for the integrated system. Experiments based on field measurements from an offshore integrated energy platform in China show that the proposed framework significantly outperforms benchmark models in terms of MAPE, RMSE and MAE. The results demonstrate superior predictive accuracy, robustness, and degree of automation.
\end{abstract}

\begin{highlights}
\item MVMD extracts multi-scale features for integrated wind–solar–wave forecasting.

\item Bayesian optimization auto-tunes MVMD parameters for higher accuracy.

\item MVMD–LSTM improves temporal feature extraction and prediction stability.

\item Validated on a real wind–solar–wave integrated platform in China.

\item Enables accurate ultra-short-term forecasts for multi-source hybrid grids.

\end{highlights}


\end{frontmatter}

\begin{table}[H]
    \caption*{\textbf{Abbreviations}}
	\renewcommand{\arraystretch}{1.1}
	\noindent
	\begin{tabular}{
			>{\raggedright\arraybackslash}p{0.3\linewidth}>{\raggedright\arraybackslash}p{0.13\linewidth}
			>{\raggedright\arraybackslash}p{0.3\linewidth}>{\raggedright\arraybackslash}p{0.13\linewidth}
		}
		\toprule
		Auto-Regressive Integrated Moving Average & ARIMA & Variational Mode Decomposition & VMD \\
		Artificial Neural Network & ANN &  Multivariate Variational Mode Decomposition & MVMD \\
		Support Vector Regression & SVR & Intrinsic Mode Functions & IMFs \\
		Random Forests & RF & Alternating Direction Method of Multipliers & ADMM \\
		Convolutional Neural Network & CNN & Mean Absolute Percentage Error & MAPE \\
		Residual Network & ResNet & Root Mean Square Error & RMSE \\
		Long Short-Term Memory & LSTM & Mean Absolute Error & MAE \\
		\bottomrule
	\end{tabular}
\end{table}

\section{Introduction}

Currently, the global energy structure is rapidly transitioning towards cleaner and lower-carbon, with the proportion of clean energy in energy consumption continuously increasing \cite{tian2022global,hassan2024renewable}. Against this backdrop, renewable energy sources such as wind, solar and wave power are being extensively developed and utilized due to their abundant resources and environmentally friendly characteristics \cite{widen2015variability}. However, the power output of a single renewable energy source is greatly affected by natural variability and its instability severely limits large-scale grid integration \cite{che2025impact}. To address this challenge, complementary multi-energy power generation systems—integrating wind, solar, wave and other renewable energy sources—should be developed to enable coordinated operation that smooths power output, improves energy utilization efficiency, and promotes the green transformation of power systems \cite{jurasz2020review}. To validate the feasibility of coordinated multi-energy generation in real-world settings, a marine wind–solar–wave integrated power generation platform was deployed on Wanshan Island, Zhuhai, China. The device incorporates 62.8 kW of solar energy, 60 kW of wave energy, and 6 kW of wind power, enabling three-source coupling and complementary generation on a single platform. The natural differences in output characteristics—solar radiation, sea state, and wind variability—make these sources mutually complementary but also dynamically interrelated. These complex interactions introduce significant uncertainty in power output, making accurate forecasting of integrated wind–solar–wave systems essential for stable and efficient operation. Through coordinated multi-energy generation, these systems can not only smooth power output and improve energy utilization efficiency, but can also effectively promote the intelligent and green upgrading of power systems \cite{modu2023systematic,khalid2024smart}.

With the widespread deployment of multi-energy complementary generation systems, issues related to their operation management and optimal scheduling have become increasingly prominent \cite{wei2024optimization}. As a fundamental component for the safe and economic operation of power systems, power forecasting plays an important role in load balancing \cite{hasan2025state}, equipment dispatching \cite{wang2024economics} and renewable energy integration \cite{ahmed2019review}. In particular, within multi-energy complementary systems, the diverse resource characteristics and complex coupling mechanisms of different energy sources lead to more varied and difficult-to-model temporal changes in power output \cite{wang2024multi}. Therefore, achieving high-precision forecasting of power generation in multi-energy complementary systems has become an urgent issue in the field of renewable energy, with significant practical importance for the reliable operation and scientific planning of power systems.

The power output of wind-solar-wave power can be categorized according to the forecasting horizon into ultra-short-term (a few seconds to 4 hours), short-term (4 hours to one day), medium-term (one day to one week) and long-term forecasting (beyond one week), and this classification has been widely adopted in the field of renewable energy forecasting \cite{hong2020energy,hong2019ultra}. Ultra-short-term forecasts provide essential information for decision-makers to optimize load tracking and device control; short-term forecasts are mainly used for pre-dispatch allocation, while medium-term and long-term forecasts are primarily applied in maintenance planning \cite{jung2014current}. With the increasing penetration of wind, solar and wave power in power grids, the demand for higher accuracy in ultra-short-term forecasting of wind-solar-wave power output within a few minutes has become more urgent \cite{anvari2016short}. Accurate ultra-short-term power forecasts can effectively mitigate the impact of power fluctuations on grid stability and assist system operators in timely scheduling adjustments \cite{yu2022ultra}. Due to the complex resource characteristics and coupling mechanisms of multi-energy complementary systems, grid operators increasingly rely on more precise forecasts to ensure safe and stable operation of the power grid \cite{lafuente2025state}. Furthermore, compared to other forecasting horizons, ultra-short-term forecasting of wind-solar-wave generation requires not only higher accuracy, but is also significantly more challenging due to the much shorter time window \cite{yu2024ultra}. Therefore, this study aims to enhance the accuracy of ultra-short-term forecasting for wind-solar-wave power generation.

Accurate forecasting of wind, solar and wave power generation has had significant economic impacts and technical advantages \cite{hassan2023review,guo2021review}. As a result, researchers have developed and refined a variety of forecasting methods for wind power, solar power and wave power. These include traditional approaches such as the Auto-Regressive Integrated Moving Average (ARIMA) model \cite{alsamamra2024performance,chodakowska2023arima,reikard2009forecasting}, as well as machine learning methods including Artificial Neural Network (ANN) \cite{atecs2023estimation,ding2011ann,hadadpour2014wave}, Support Vector Regression (SVR) \cite{yuan2022wind,das2017svr}, and Random Forests (RF) \cite{zhou2016wind,guan2024study}. At the same time, deep learning methods such as Convolutional Neural Network (CNN) \cite{qu2021day}, Residual Network (ResNet) \cite{mirza2023hybrid}, and Long Short-Term Memory (LSTM) are also widely used in power prediction applications. As Long Short-Term Memory (LSTM) has gained popularity in power forecasting applications due to their superior capability in modeling complex nonlinear data and effectively capturing temporal dependencies in time series \cite{zhang1998forecasting}. Zhang et al.(2019) applied LSTM to wind power prediction and demonstrated that it outperformed traditional methods such as Support Vector Machines in forecast accuracy \cite{zhang2019wind}. Similarly, According to Konstantinou et al.(2021), a stacked LSTM model was used to forecast the power output of a PV plant in Nicosia, Cyprus, showing superior accuracy compared to ARIMA, ANN and other conventional approaches \cite{konstantinou2021solar}. Minuzzi et al.(2023) employed LSTM models for wave power prediction, effectively capturing the sequential characteristics of wave power and achieving accurate forecasts \cite{minuzzi2023deep}. Additionally, Chen et al.(2023) investigated the combined forecasting of PV and wind power using LSTM, demonstrating enhanced overall prediction performance compared to individual models \cite{chen2023regional}. The ability to jointly forecast wind, solar and wave power is of practical importance for dispatch and planning in smart grids. Nevertheless, existing research predominantly targets single-source forecasting or wind–PV joint models;a unified and comprehensive framework that jointly forecasts wind, solar and wave power remains lacking. Addressing this gap, this study explores ultra-short-term power forecasting methods from the combined perspective of wind, solar and wave power.

Due to the intermittent characteristics of renewable energy sources such as wind, solar and wave power, multi-energy power data exhibit high randomness and volatility \cite{sinsel2020challenges}. Therefore, relying solely on a single forecasting method often fails to achieve accurate power prediction results. To address this issue, some researchers have proposed hybrid approaches that first perform signal decomposition and then train deep learning predictors \cite{qian2019review}. Currently, decomposition-based methods have become one of the most widely used data preprocessing techniques and have demonstrated promising forecast performance. In decomposition-based hybrid methods, the original data are first decomposed into several relatively stationary subsequences using decomposition techniques. Then, individual forecasting models are developed for each subsequence. Finally, the forecasting results of all subsequences are aggregated to obtain the final prediction. By decomposing data into multiple relatively stationary subsequences and predicting them separately, this approach can effectively improve the accuracy of power forecasting. Variational Mode Decomposition (VMD) is a recently popular signal decomposition method that can decompose nonstationarity raw data into a set of narrowband intrinsic mode functions with distinct frequency characteristics, thus significantly enhancing the performance of forecasting models \cite{dragomiretskiy2013variational,wu2022completed,gu2022review}. However, traditional VMD primarily targets univariate data and cannot fully exploit the coupling characteristics among multi-energy data \cite{ghanbari2025short}. Therefore, this study introduces the Multivariate Variational Mode Decomposition (MVMD) method, which can simultaneously process multiple correlated variables during decomposition, more effectively preserving the correlations among multi-source information, thus further improving the accuracy and stability of power forecasting \cite{ur2019multivariate}.

Although progress has been made, several challenges remain in multi-source renewable power forecasting. Firstly, integrated wind-solar-wave power output sequences are highly nonlinear and nonstationarity, harming forecast model accuracy and generalization. Secondly, the performance of signal decomposition methods highly depends on the selection of decomposition parameters \cite{eriksen2022data}. The most existing studies rely on manual or empirical parameter tuning, making the models difficult to adapt to changes in data characteristics and limiting their practicality and automation level \cite{zulfiqar2023hybrid}. Therefore, it is necessary to develop new multi-energy complementary system power forecasting methods that can automatically optimize decomposition parameters and enhance the capability of time series feature extraction. Based on the above questions, we introduce Bayesian optimization, which is an effective and automated strategy for tuning decomposition parameters and enables models to adapt varying data conditions without manual intervention. 

Based on the above, this study proposes an adaptive parameter-optimized Multivariate Variational Mode Decomposition–Long Short-Term Memory network (MVMD-LSTM) method for integrated wind, solar and wave power forecasting. Specifically, the proposed method consists of the following. Firstly, to fully capture the coupling characteristics among wind, solar and wave power signals, MVMD is employed to jointly decompose multiple correlated time series into a set of Intrinsic Mode Functions (IMFs). This enables the extraction of more representative features that reflect the dynamic interactions among multi-source data.Then, to ensure optimal decomposition performance, Bayesian optimization is integrated to adaptively select MVMD parameters, such as the number of modes and the penalty factor. This automatic tuning process improves the decomposition quality. Lastly, the decomposed IMFs are used as inputs to LSTM, which is capable of modeling complex temporal dependencies and nonlinear patterns in the multi-energy data. This facilitates accurate ultra-short-term forecasting of the integrated power output.

The main contributions of this study are summarized as follows. Firstly, this study integrates wind, solar and wave power into a unified forecasting framework, effectively capturing their complementary characteristics and coupling relationships to improve multi-source renewable energy prediction. Secondly, a Bayesian-optimized MVMD-LSTM model is developed, where Bayesian optimization adaptively determines decomposition parameters, enhancing accuracy, stability, and automation compared to conventional tuning approaches. Lastly, experiments on a real-world marine multi-energy datasets show that the proposed method consistently outperforms benchmark models across multiple metrics, achieving superior forecasting accuracy and robustness.

The remainder of this paper is organized as follows. Section \ref{sec:background} briefly reviews the background theory of our proposed method. Section \ref{sec:framwork} describes the framework of our proposed method. Section \ref{sec:experimental} conducts sufficient experiment on datasets. Section \ref{sec:Discussion} focuses on the analysis and interpretation of the results, while Section \ref{sec:conclusion} summarizes the contributions of the study and outlines its limitations and future research directions.

\section{Preliminary Knowledge}
\label{sec:background}

In this section, the preliminary knowledge of the methods involved in our proposed method are described in detail.

\subsection{Multivariate Variational Mode Decomposition (MVMD)}

Let the original three-channel signal, comprising the wind, solar and wave power sequences, be denoted as $\{x_j(t)\}_{j=1}^3$. Given a prescribed number of decomposition modes \(K\), we seek to extract the intrinsic mode functions \(\{\{u_{j,k}(t)\}_{j=1}^3\}_{k=1}^K\). 

Construct the constrained multivariate variational optimization problem of MVMD. The goal of this problem is to minimize the sum of the bandwidths of the extracted IMFs while ensuring that the sum of the extracted IMFs can exactly recover the original signal. The constrained optimization problem is given by the following equation:
\begin{equation}\label{eq:mvmd_opt}
\begin{aligned}
&\min_{\{u_{j,k}\},\{\omega_k\}}
\sum_{k=1}^K\sum_{j=1}^3
\left\|\partial_t\bigl[u_{j,k}^+(t)\,e^{-\mathrm{i}\,\omega_k t}\bigr]\right\|_2^2\\
&\text{s.t.}\quad
\sum_{k=1}^K u_{j,k}(t) = x_j(t),\quad j=1,2,3
\end{aligned}
\end{equation}

where $u_{j,k}(t)$ is the $k$-th intrinsic mode function (IMF) of the $j$-th channel signal at time $t$; $\omega_k$ is the center (angular) frequency associated with the $k$-th IMF, shared across all channels; $K$ is the total number of IMFs to be extracted; $j=1,2,3$ indexes the three input channels; $x_j(t)$ is the original $j$-th channel signal at time $t$; $u_{j,k}^+(t)$ denotes the analytic signal of $u_{j,k}(t)$ obtained via the Hilbert transform; $\partial_t$ is the partial derivative operator with respect to time $t$; $\|\cdot\|_2$ denotes the $L^2$ (Euclidean) norm; $\mathrm{i}$ is the imaginary unit satisfying $\mathrm{i}^2=-1$; and $t$ is the continuous time variable.

Construct an augmented Lagrangian function to remove the constraint of the multivariate variational optimization problem. The augmented Lagrangian function contains two penalty terms: a quadratic term \(\displaystyle \sum_{j=1}^3 \bigl\|x_j(t) - \sum_{k=1}^K u_{j,k}(t)\bigr\|_2^2\) to ensure the reconstruction accuracy and a Lagrangian multipliers term \(\displaystyle \sum_{j=1}^3 \bigl\langle \lambda_j(t),\,x_j(t) - \sum_{k=1}^K u_{j,k}(t)\bigr\rangle\) to ensure that the constraints are strictly satisfied. The Equation of the 
augmented Lagrangian function is as follows:

\begin{equation}\label{eq:aug_lag}
\begin{split}
\mathcal{L}(\{u_{j,k}\},\{\omega_k\},\{\lambda_j\})
&= \alpha \sum_{k=1}^K \sum_{j=1}^3
  \Bigl\|\partial_t\bigl[u_{j,k}^+(t)e^{-\mathrm{i}\omega_k t}\bigr]\Bigr\|_2^2 \\
&\quad + \sum_{j=1}^3 \Bigl\|x_j(t) - \sum_{k=1}^K u_{j,k}(t)\Bigr\|_2^2 \\
&\quad + \sum_{j=1}^3 \bigl\langle \lambda_j(t),\,x_j(t) - \sum_{k=1}^K u_{j,k}(t)\bigr\rangle
\end{split}
\end{equation}

Solve the variational optimization problem in Eq. (\ref{eq:aug_lag}) by Alternating Direction Method of Multipliers (ADMM) \cite{tang2023novel}. The advantage of ADMM is that it can convert the entire complex optimization problem into a set of sub-optimization problems, thereby reducing the difficulty of the solution process. The update Equations for the mode $\hat u_{j,k}(\omega)$, center frequency $\omega_k$, and Lagrangian multipliers $\lambda_j^{(t)}$ are shown in Eq. (\ref{eq:admm_u}), (\ref{eq:admm_omega}), (\ref{eq:admm_lambda}).

\begin{align}
\hat u_{j,k}^{(n+1)}(\omega)
&=\frac{\hat x_j(\omega)
 - \sum_{i<k}\hat u_{j,i}^{(n+1)}(\omega)
 - \sum_{i>k}\hat u_{j,i}^{(n)}(\omega)
 + \tfrac12\,\hat\lambda_j^{(n)}(\omega)}
 {1 + 2\alpha\,(\omega-\omega_k^{(n)})^2} \label{eq:admm_u}\\
\omega_k^{(n+1)}
&=\frac{\displaystyle\sum_{j=1}^3\int_0^\infty 
   \omega\,\bigl|\hat u_{j,k}^{(n+1)}(\omega)\bigr|^2\,d\omega}
 {\displaystyle\sum_{j=1}^3\int_0^\infty
   \bigl|\hat u_{j,k}^{(n+1)}(\omega)\bigr|^2\,d\omega} \label{eq:admm_omega}\\
\hat\lambda_j^{(n+1)}(\omega)
&=\hat\lambda_j^{(n)}(\omega)
 + \tau\Bigl(\hat x_j(\omega)
   - \sum_{k=1}^K \hat u_{j,k}^{(n+1)}(\omega)\Bigr) \label{eq:admm_lambda}
\end{align}

where $\hat x(\omega)$, $\hat\lambda(\omega)$ and $\hat u(\omega)$ are the Fourier transforms of $x(t)$, $\lambda(t)$ and $u(t)$.

\begin{figure}[H]
	\centering
	\includegraphics[width=0.75\textwidth]{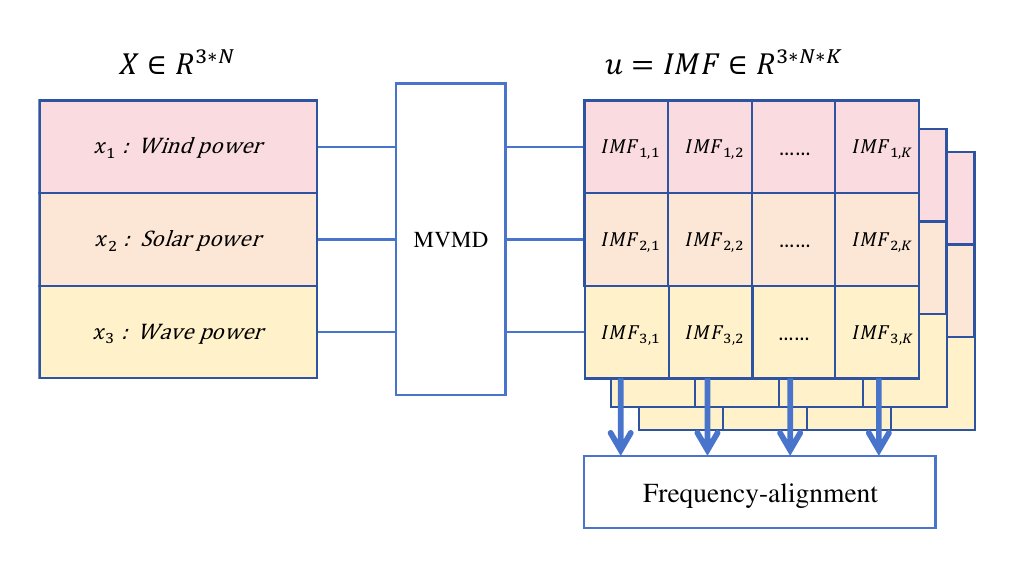}
	\caption{The configuration of MVMD cell.}
	\label{fig:mvmd_structure}
\end{figure}

Finally, after several iterations, the wind, solar and wave power sequences are decomposed into IMFs $\{x_j(t)\}_{j=1}^3 \in \mathbb{R}^{N\times 3}$ and corresponding center frequencies $\{\omega_k\}_{k=1}^K \in \mathbb{R}^{K\times 1}$, where N represents the sequence length. The configuration of the MVMD cell is illustrated in \cref{fig:mvmd_structure}.

\subsection{Long Short-Term Memory (LSTM)}
\begin{figure}[H]
    \centering
    \includegraphics[width=0.75\textwidth]{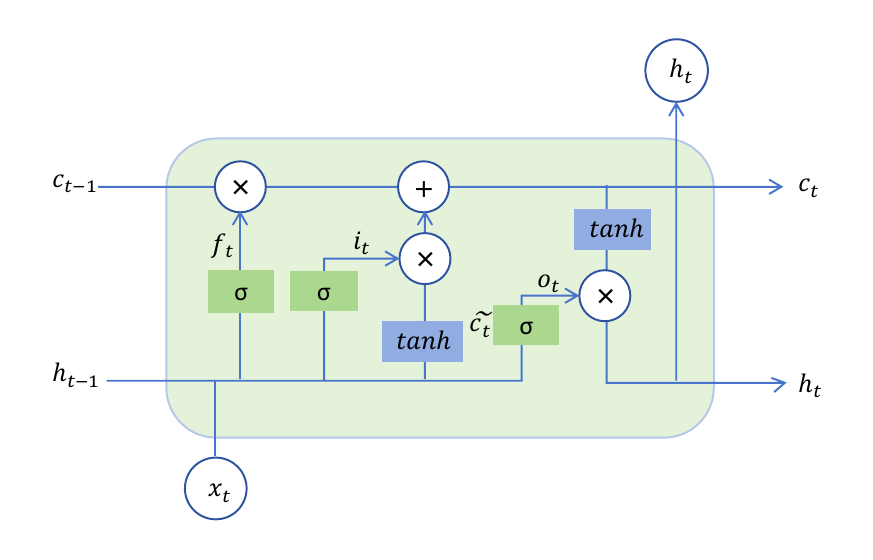}
    \caption{The configuration of LSTM cell.}
    \label{fig:lstm_structure}
\end{figure}

LSTM is a prominent deep learning method for time series forecasting. It has an excellent memory capability and it can find regular information from power historical data. LSTM introduces the “gates” mechanism to enhance the basic functionalities of recurrent cell memory. This gate mechanism enables LSTM to control the flow of information. Thus, LSTM is able to preserve important information for a longer period and ignore less useful historical information from time series data. Due to its special architecture, LSTM is suitable for ultra-short-term wind, solar and wave power forecasting. \cref{fig:lstm_structure} illustrates the structure of the LSTM cell which is comprised of an input gate, output gate, and forget gate. 

The equations for the LSTM model are as follows:
\begin{align}
    f_t &= \sigma(W_f [h_{t-1}, x_t] + b_f) \\
    i_t &= \sigma(W_i [h_{t-1}, x_t] + b_i) \\
    o_t &= \sigma(W_o [h_{t-1}, x_t] + b_o) \\
    \tilde{c}_t &= \tanh(W_c [h_{t-1}, x_t] + b_c) \\
    c_t &= f_t \otimes c_{t-1} + i_t \otimes \tilde{c}_t \\
    h_t &= o_t \otimes \tanh(c_t)
\end{align}
\noindent
where $W_f, W_i, W_c, W_o$ are the sets of weights, $b_f, b_i, b_c, b_o$ are the corresponding bias vectors, and $\otimes$ denotes element-wise multiplication. $\sigma$ is the sigmoid function defined as $\sigma(x) = \frac{1}{1 + e^{-x}}$, and tanh is the hyperbolic tangent function defined as $\tanh(x) = \frac{e^{x} - e^{-x}}{e^{x} + e^{-x}}$. $f_t$ is the forget gate, $i_t$ is the input gate, $o_t$ is the output gate, $\tilde{c}_t$ is the candidate cell state, $c_t$ is the cell state at time step $t$, $h_t$ is the output of the LSTM at time $t$, and $x_t$ is the input. The forget gate determines what information should be removed and the input gate decides what information should be kept.

\subsection{Bayesian Optimization}

In this study, Bayesian optimization is embedded into the MVMD component to automate its parameter configuration. Bayesian optimization is a powerful global optimization technique specifically designed for expensive, gradient-free, potentially noisy or highly nonlinear objective functions \cite{shahriari2015taking,frazier2018tutorial}. Unlike traditional hyperparameter optimization approaches such as grid search or random search, Bayesian optimization builds a probabilistic surrogate model—commonly a Gaussian process—to model the objective function \cite{snoek2012practical}. This surrogate model is iteratively updated with previous observations, and an acquisition function determines the next point to evaluate, balancing exploration of uncertain regions and exploitation of known optimal regions.

The general procedure of Bayesian optimization involves three steps \cite{wu2017bayesian}. Firstly, fitting a probabilistic surrogate model to approximate the objective function. Then using an acquisition function such as Expected Improvement, Knowledge Gradient, or Entropy Search to select the next evaluation point. In this study, we use the Expected Improvement to choose the next evaluation point. Lastly updating the surrogate model with new observations and repeating the process until a stopping criterion is met.

Formally, Bayesian optimization aims to solve
\begin{equation}
x^* = \arg\min_{x\in X} f(x)
\end{equation}
where \( f(x) \) is the objective function, which is the prediction error MAPE in a validation set and \( X \) denotes the hyperparameter space.

Bayesian optimization hyperparameter tuning methods have attracted much attention, achieving near-optimal solutions with fewer evaluations than manual tuning or grid search \cite{wang2016bayesian}. In this study, Bayesian optimization is used to automatically determine the optimal decomposition number \( K \) and penalty parameter \( \alpha \) for the MVMD module within the proposed hybrid MVMD-LSTM forecasting framework \cite{habtemariam2023bayesian}. The objective function is defined as the prediction error MAPE in the validation and Bayesian optimization sequentially selects parameter combinations to minimize this error. Compared to manual tuning or grid search, this strategy offers higher search efficiency, better global optima through uncertainty-guided exploration and reduced computational cost.

\section{The framework of our proposed method}
\label{sec:framwork}
\begin{figure}[H]
	\centering
	\includegraphics[width=0.75\textwidth]{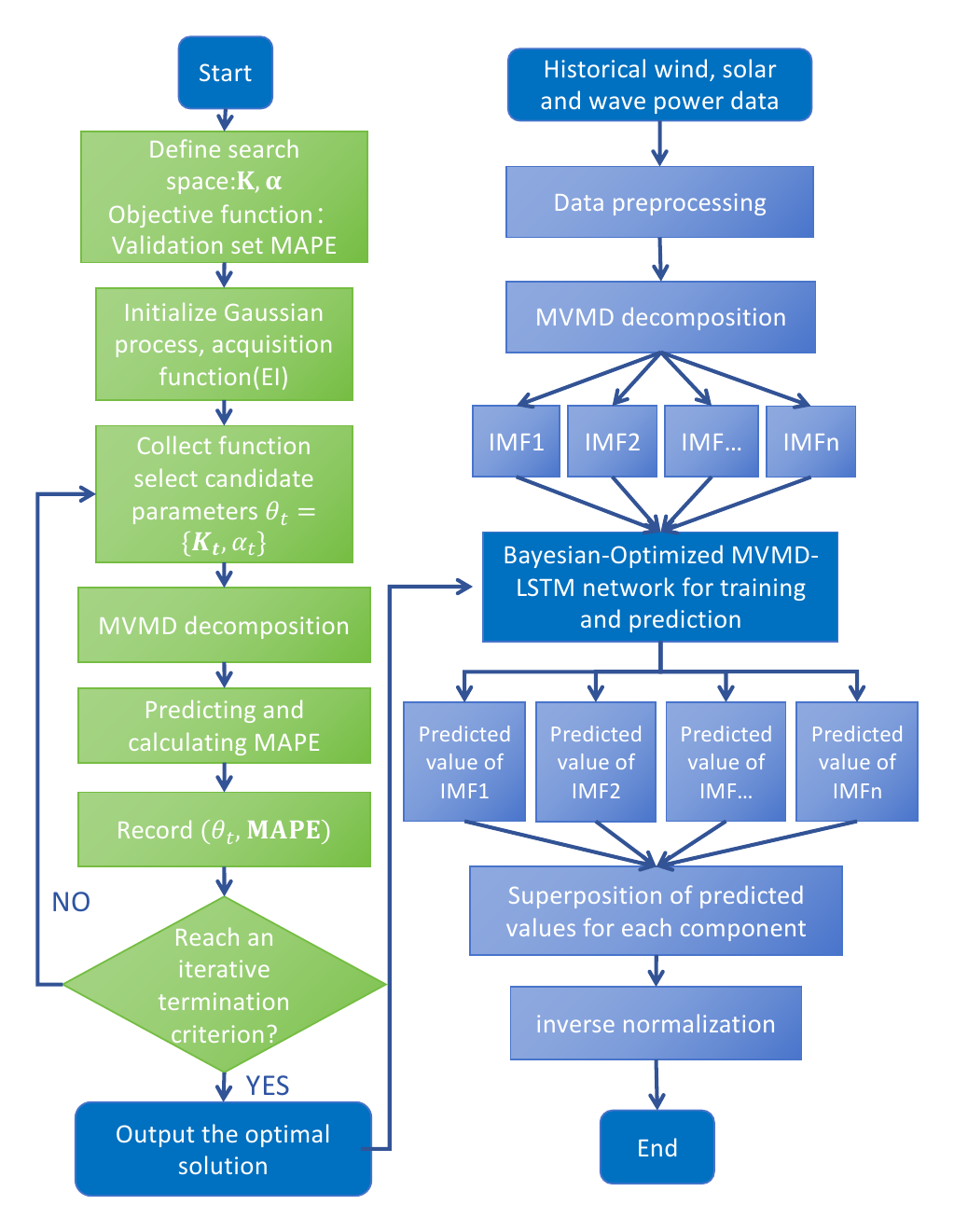}
	\caption{Bayesian-Optimized MVMD-LSTM prediction flowchart.}
	\label{fig:prediction flowchart}
\end{figure}

\begin{figure}[H]
    \centering
    \includegraphics[width=0.75\textwidth]{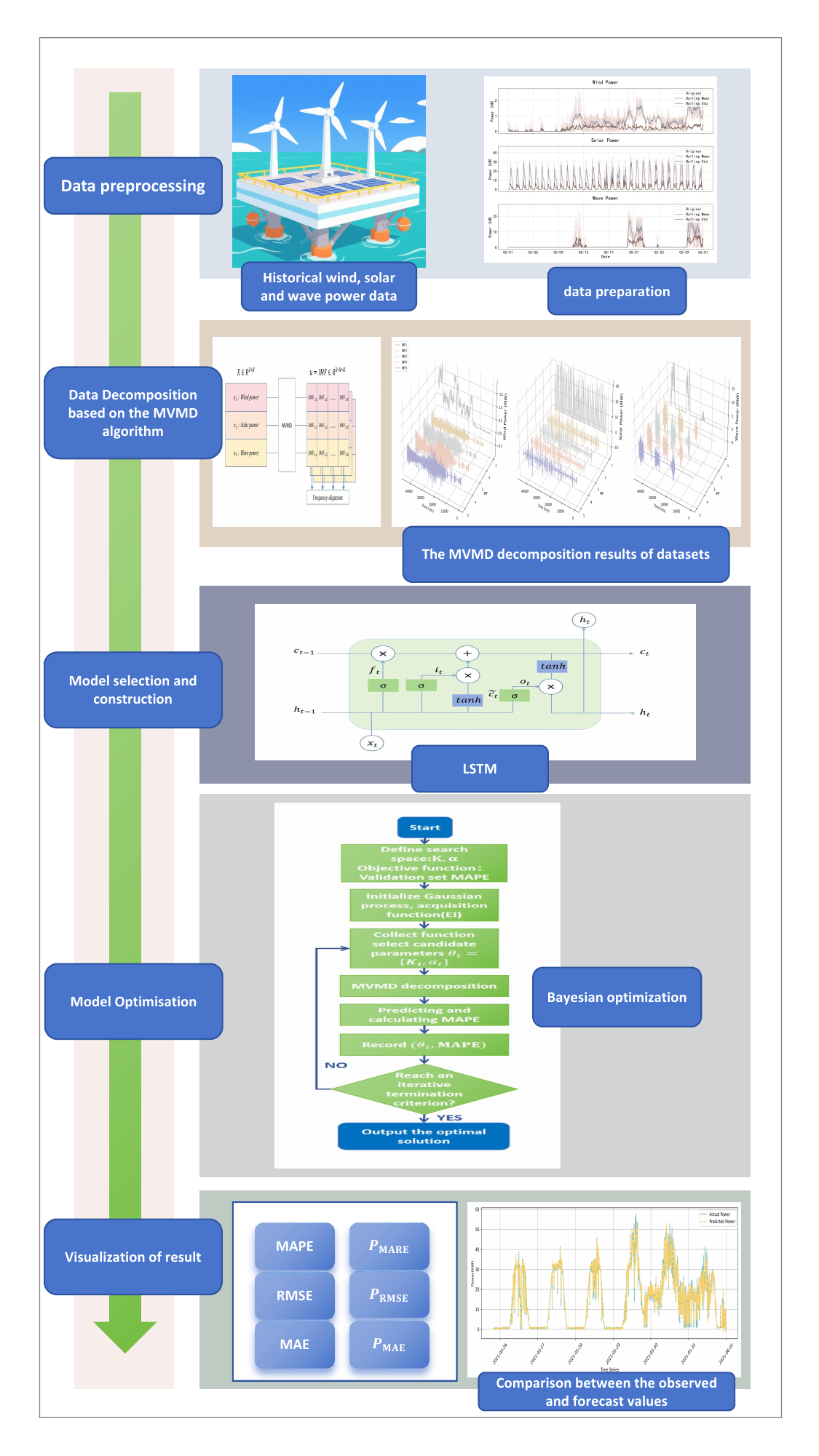}
    \caption{Framework of the prediction model proposed in this paper.}
    \label{fig:Framework of the prediction model proposed in this paper}
\end{figure}

The flow of the proposed Bayesian-Optimized MVMD-LSTM for integrated wind, solar and wave power prediction method is shown in \cref{fig:prediction flowchart}, which mainly includes five stages: data preprocessing, data decomposition, model construction, parameter optimization, and evaluation and comparison of observed and forecast values (see \cref{fig:Framework of the prediction model proposed in this paper}).

Phase 1: Data preprocessing. The collected historical power generation data are analyzed to identify their characteristics, followed by basic preprocessing.

Phase 2: Data decomposition. The original data are decomposed using MVMD to mitigate nonlinearity and nonstationarity, thereby producing a set of more stationary and regular subsequences.

Phase 3: Model construction. An LSTM model is constructed using the MVMD data as inputs to forecast wind, solar and wave power, aiming to improve predictive accuracy.

Phase 4: Parameter optimization. The model hyperparameters are optimized using Bayesian optimization, and the resulting optimal values are then applied to the model to improve prediction performance.

Phase 5: Model evaluation. A series of prediction metrics are employed to evaluate the proposed and benchmark models, with actual performance results used to validate the predictive accuracy of the proposed algorithm.

\section{Experimental result}
\label{sec:experimental}
\subsection{Dateset}
\begin{figure}[H]
	\centering
	\includegraphics[width=0.75\textwidth]{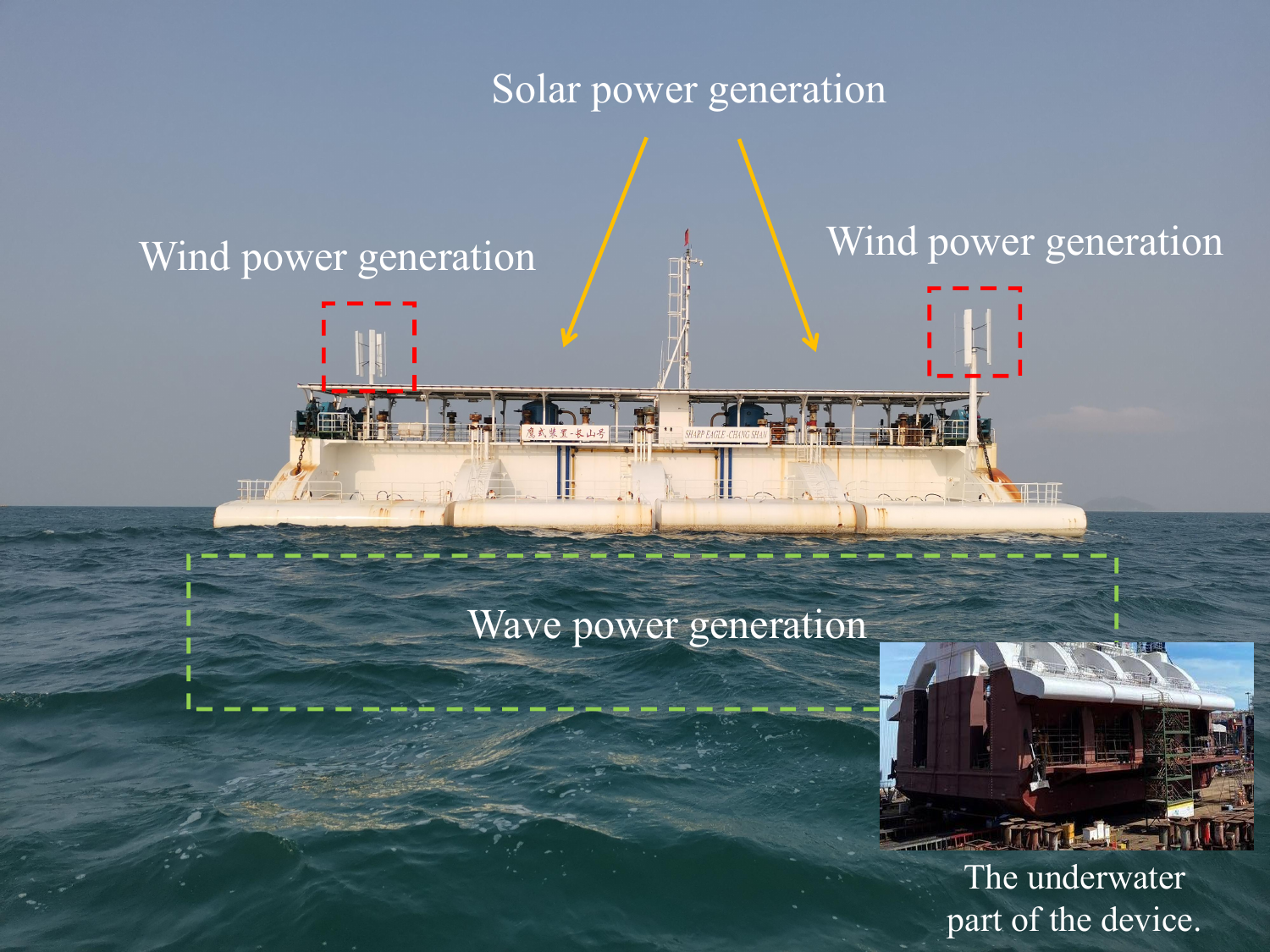}
	\caption{The integrated wind-solar-wave power generation platform.}
	\label{fig:equip}
\end{figure}

\begin{figure}[H]
	\centering
	\includegraphics[width=0.75\textwidth]{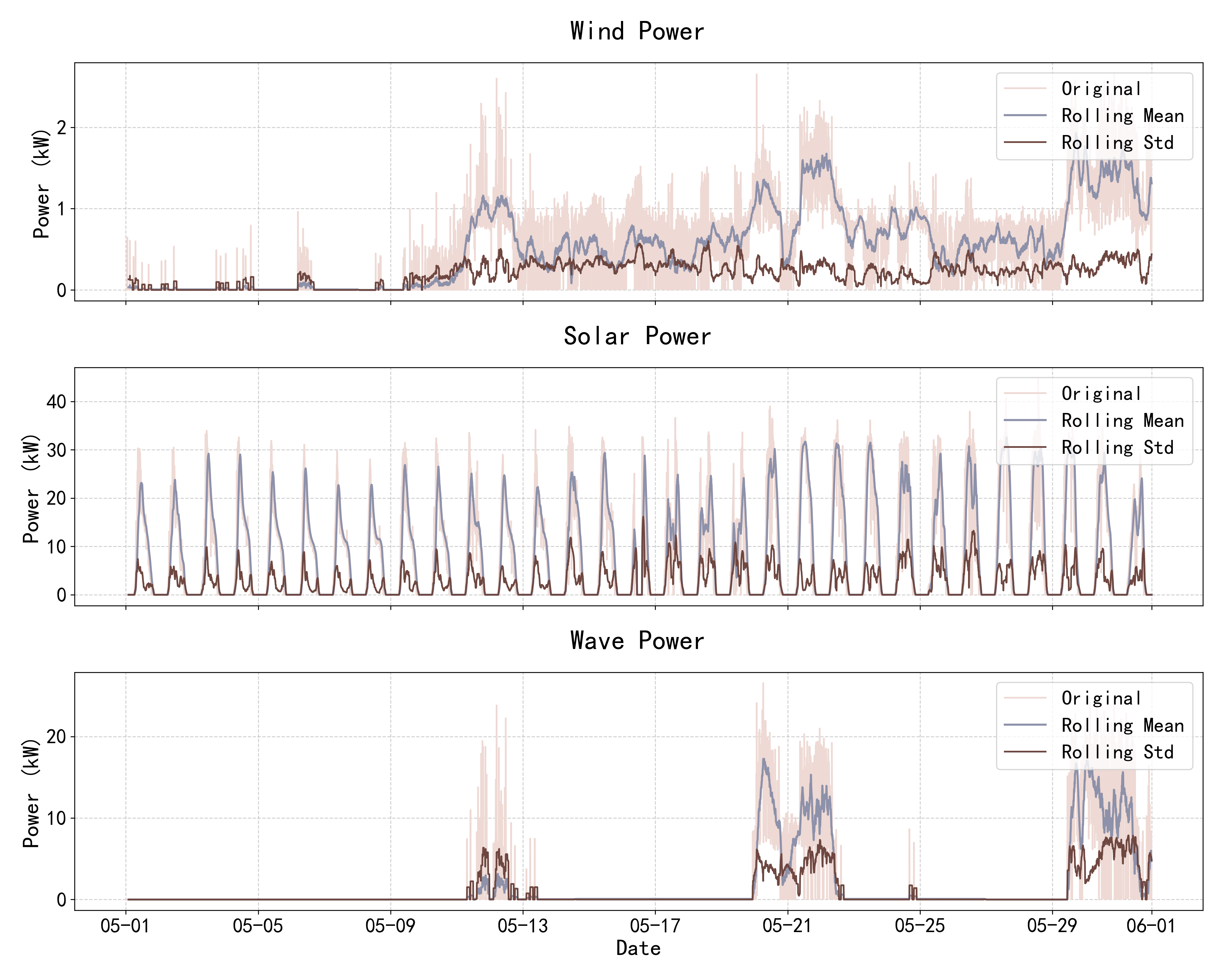}
	\caption{Nonstationarity Analysis of Renewable Power Time Series in May.}
	\label{fig:May_Power_Nonstationarity}
\end{figure}

\begin{figure}[H]
	\centering
	\includegraphics[width=0.75\textwidth]{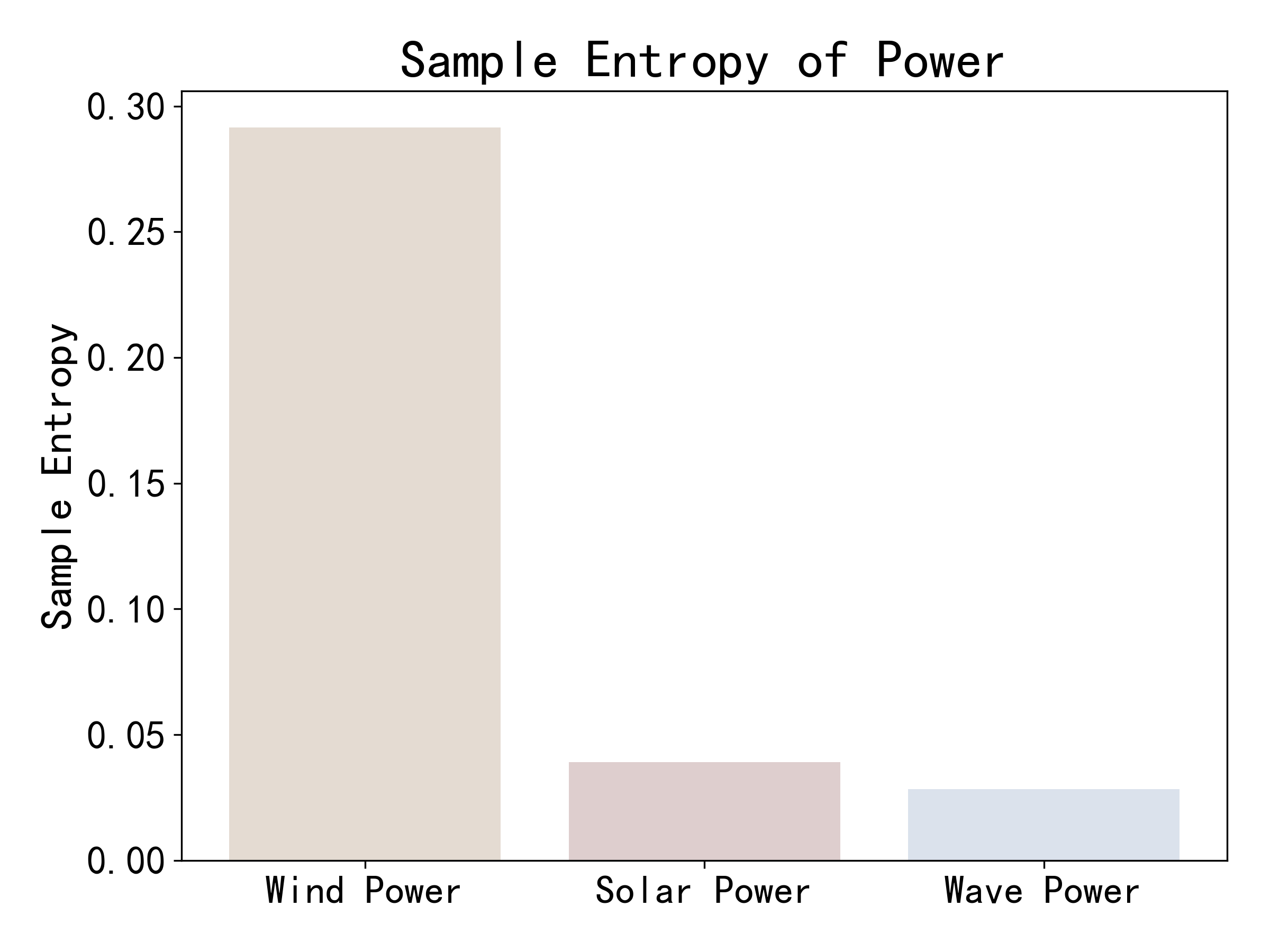}
	\caption{Nonlinear Analysis of Renewable Power Time Series in May.}
	\label{fig:May_Sample_Entropy}
\end{figure}

The datasets used in this study comes from a floating multi-energy complementary platform that integrates wind, solar and wave power generation on Wanshan Island, Zhuhai, Guangdong \cite{guangzhou2023offshore}. The installation has a total capacity of 128.8 kW (wave: 60 kW; solar: 62.8 kW; wind: 6 kW), and its equip is shown in \cref{fig:equip}. We use 5-minute observations from May to September 2021, splitting the data 80/20 into training and test sets. This real-world datasets supports analysis of cross-source temporal interactions and complementarity under marine conditions and is used to validate the proposed method. As shown in Fig.\ref{fig:May_Power_Nonstationarity}, the raw power series and their rolling statistics exhibit pronounced nonstationarity. Moreover, Fig.\ref{fig:May_Sample_Entropy} shows that wind power has the highest sample entropy, indicating the greatest structural complexity, whereas solar and wave power present lower but non-negligible nonlinearity. These nonstationary and nonlinear characteristics complicate accurate power forecasting.

\subsection{Experimental setting}
Experiments were performed on a PC equipped with an Intel Core i5-8265U processor (1.60 GHz) and 8.00 GB of RAM using Python 3.7. The parameters $K$ and $\alpha$ of MVMD were adaptively tuned through Bayesian optimization implemented with the Optuna framework \cite{akiba2019optuna}, with the MVMD algorithm based on the publicly available MVMD.py script on GitHub \cite{Dmocrito2024mvmd}. The LSTM module \cite{hochreiter1997long} was developed using the TensorFlow/Keras framework and trained using the Adam optimizer \cite{kinga2015method} to balance training efficiency and stability. The LSTM learning rate was set to 0.001 and the model was trained for 100 epochs.

To identify the optimal number of hidden neurons and batch size for the proposed method, preliminary experiments were conducted. Various combinations of neuron counts (32, 64, and 128) and batch sizes were evaluated, revealing that 32 neurons with a batch size of 64 achieved the best performance. Accordingly, this configuration was adopted for model training. In this study, a 5-minute-ahead forecast (\(X_t\)) was performed, using the past half-hour of data (\(X_{t-1}\) to \(X_{t-6}\)) as input to the prediction model.

\subsection{Evaluation metrics}
In this study, three common evaluation metrics were used, which are Mean Absolute Percentage Error (MAPE), Root Mean Square Error (RMSE) and Mean Absolute Error (MAE) \cite{karijadi2023wind}. The three metrics are defined as follows:
\begin{equation}
	\mathrm{MAPE} = \frac{1}{N} \sum_{i=1}^{N} \left| \frac{ \hat{y}_i - y_i}{y_{\max}} \right|
\end{equation}

\begin{equation}
	\mathrm{RMSE} = \sqrt{ \frac{1}{N} \sum_{i=1}^{N} \left( \hat{y}_i - y_i \right)^2 }
\end{equation}

\begin{equation}
	\mathrm{MAE} = \frac{1}{N} \sum_{i=1}^{N} \left| \hat{y}_i - y_i \right|
\end{equation}

where $y_i$ is the real power value at time $i$ and $\hat{y}_i$ is the predicted power value at time $i$. $N$ denotes the number of data points and $y_{\max}$ is the maximum value of the whole data. The lower the value of the evaluation metrics above, the better the performance of the forecasting method. The percentage of improvement metric is also applied in this study to evaluate the enhancement of the proposed forecasting method compared to other forecasting methods \cite{liu2020new}. The percentage of improvements are calculated as follows:

\begin{equation}
	\mathrm{P}_{\mathrm{RMSE}} = \frac{\mathrm{RMSE}_1 - \mathrm{RMSE}_2}{\mathrm{RMSE}_1} \times 100\%
\end{equation}

\begin{equation}
	\mathrm{P}_{\mathrm{MAPE}} = \frac{\mathrm{MAPE}_1 - \mathrm{MAPE}_2}{\mathrm{MAPE}_1} \times 100\%
\end{equation}

\begin{equation}
	\mathrm{P}_{\mathrm{MAE}} = \frac{\mathrm{MAE}_1 - \mathrm{MAE}_2}{\mathrm{MAE}_1} \times 100\%
\end{equation}

\subsection{Results}
\begin{figure}[H]
	\centering
	\includegraphics[width=0.75\textwidth]{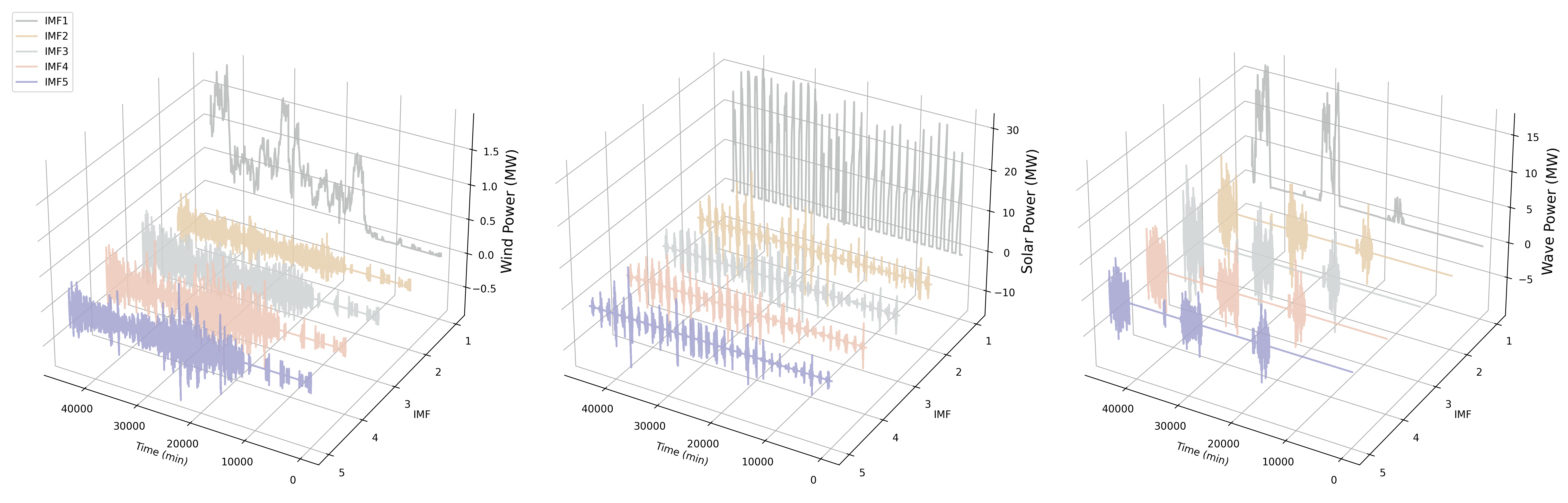}
	\caption{The MVMD decomposition results of datasets.}
	\label{fig:MVMD-decomposition}
\end{figure}

In the initial stage of the proposed method, the original power data are decomposed into multiple subseries using the MVMD. As the nonlinearity and nonstationarity of the wind, solar and wave power data, MVMD is employed to extract a set of IMFs that exhibit lower degrees of nonlinearity and nonstationarity. As shown in \cref{fig:MVMD-decomposition}, the wind, solar and wave power outputs are decomposed into a set of IMFs with distinct frequency characteristics. For wind power, high-frequency components (IMF1–IMF2) dominate, while IMF3–IMF4 capture additional low-frequency fluctuations. Solar power exhibits clear periodic patterns in IMF3–IMF5, with IMF5 showing a distinct diurnal cycle. In contrast, wave power is mainly concentrated in low-frequency components (IMF4–IMF5), with little energy in the high-frequency IMFs. These results highlight the different modal characteristics of the three energy sources.

\begin{figure}[H]
	\centering
	\includegraphics[width=0.75\textwidth]{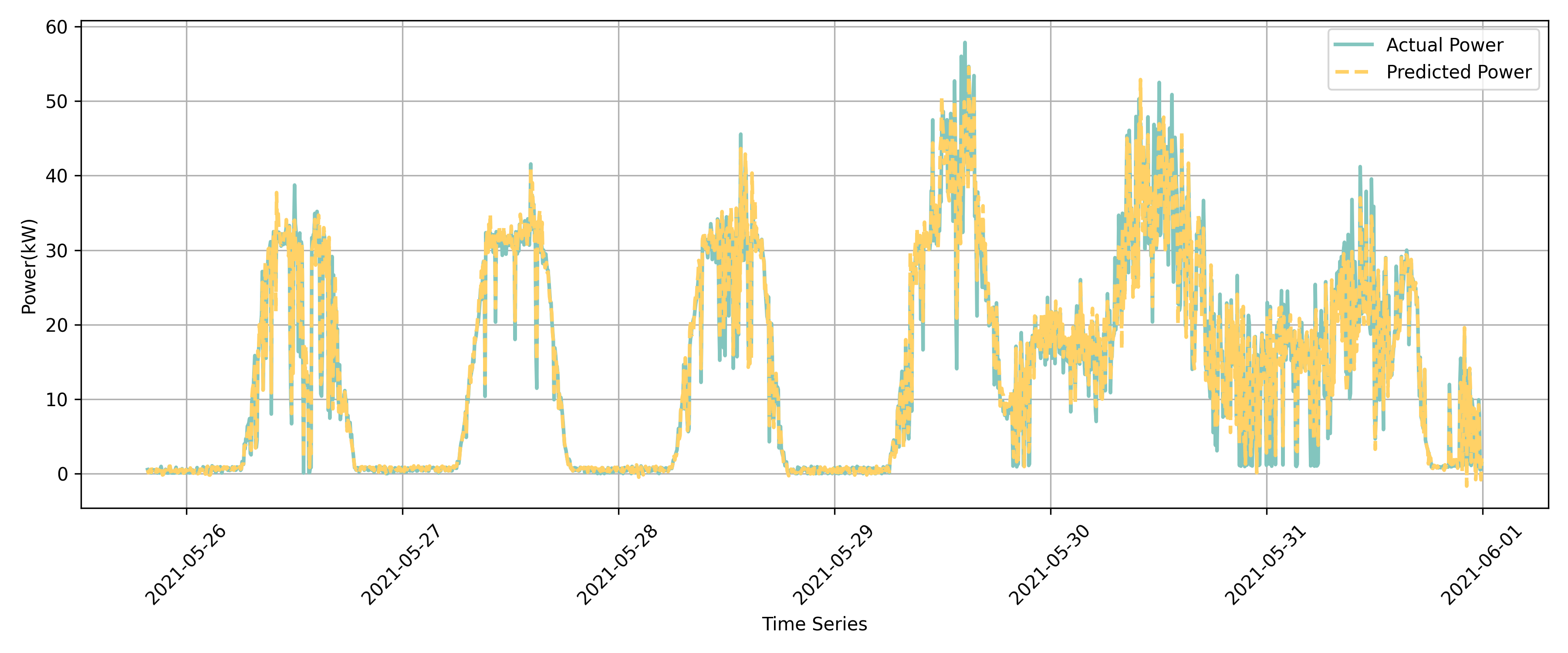}
	\caption{Comparison between the observed and forecast values.}
	\label{fig:Comparison}
\end{figure}

The LSTM model is applied to each decomposed subsequence, and the final power output is obtained by aggregating the predicted components. As shown in \cref{fig:Comparison}, the predicted curve aligns closely with the observed values, successfully capturing both the diurnal pattern and short-term fluctuations. These results demonstrate the effectiveness of the proposed method for power generation forecasting.

To evaluate the performance of the proposed Bayesian-optimized MVMD-LSTM method, seven forecasting approaches were compared: SVR, ANN, RF, CNN, ResNet, LSTM, and VMD-LSTM. The first six are standard single-model forecasting methods, while VMD-LSTM is a hybrid model that combines VMD with LSTM by decomposing the original data into subseries, forecasting each subseries individually, and aggregating the results. Table~\ref{table:metrics} displays the forecasting performance across different months, with the lowest error values highlighted in bold. The last row summarizes the average metrics across all datasets. Table~\ref{table:Percentage-improvement} reports the percentage improvements of the proposed method relative to benchmark models.The results show that LSTM consistently outperforms other single forecasting models (SVR, ANN, RF, CNN, ResNet), achieving average errors of MAPE of 4.70\%, RMSE of 4.22, and MAE of 2.70. Both hybrid methods (VMD-LSTM and the proposed method) substantially outperform all single models. On average, VMD-LSTM reduces errors to MAPE of 2.34\%, RMSE of 2.53, and MAE of 1.55, while the proposed Bayesian-optimized MVMD-LSTM further reduces them to MAPE of 1.75\%, RMSE of 1.72, and MAE of 1.16. Compared with VMD-LSTM, the proposed method achieves approximately 25\% lower MAPE, highlighting its superior accuracy.

\begin{table}[htbp]
	\centering
	\scriptsize
	\caption{Performance comparison of different forecasting methods by month and metrics.}
	\label{table:metrics}
	\begin{adjustbox}{max width=\linewidth}
		\begin{tabular}{
				>{\centering\arraybackslash}p{1.3cm}
				>{\centering\arraybackslash}p{1.3cm}
				>{\centering\arraybackslash}p{1.3cm}
				>{\centering\arraybackslash}p{1.3cm}
				>{\centering\arraybackslash}p{1.3cm}
				>{\centering\arraybackslash}p{1.3cm}
				>{\centering\arraybackslash}p{1.3cm}
				>{\centering\arraybackslash}p{1.3cm}
				>{\centering\arraybackslash}p{1.3cm}
				>{\centering\arraybackslash}p{1.3cm}
			}
			\toprule
			\multicolumn{2}{c}{} & \multicolumn{8}{c}{Forecasting Methods} \\
			\cmidrule(lr){3-10}
			Month & Metrics & SVR & ANN & RF & CNN & ResNet & LSTM & VMD-LSTM & Proposed Method \\
			\midrule
			May
			& MAPE & 5.00  & 5.09  & 4.94  & 6.25  & 5.90  & 4.89  & 3.20  & \textbf{1.68} \\
			& RMSE & 5.62  & 5.31  & 5.36  & 6.60  & 6.16  & 5.28  & 3.32  & \textbf{1.83} \\
			& MAE  & 3.31  & 3.37  & 3.28  & 4.15  & 3.91  & 3.24  & 2.12  & \textbf{1.11} \\
			Jun
			& MAPE & 6.33 & 6.53 & 6.77 & 7.62 & 8.15 & 6.50 & 3.30 & \textbf{2.84} \\
			& RMSE & 5.79 & 5.74 & 5.96 & 6.68 & 7.14 & 5.77 & 3.60 & \textbf{2.49} \\
			& MAE  & 4.20 & 4.33 & 4.49 & 5.06 & 5.40 & 4.31 & 2.19 & \textbf{1.88} \\
			Jul
			& MAPE & 4.70 & 5.23 & 4.84 & 5.49 & 6.13 & 4.91 & 2.89 & \textbf{2.46} \\
			& RMSE & 4.51 & 4.82 & 4.51 & 5.17 & 5.63 & 4.61 & 2.76 & \textbf{2.15} \\
			& MAE  & 3.12 & 3.47 & 3.21 & 3.64 & 4.07 & 3.25 & 1.91 & \textbf{1.63} \\
			Aug
			& MAPE & 1.63 & 2.35 & 1.32 & 1.81 & 1.39 & 1.14 & 0.82 & \textbf{0.76} \\
			& RMSE & 1.83 & 2.20 & 1.97 & 2.52 & 2.00 & 1.78 & 1.12 & \textbf{1.03} \\
			& MAE  & 1.08 & 1.56 & 0.88 & 1.20 & 0.92 & 0.76 & 0.54 & \textbf{0.50} \\
			Sep
			& MAPE & 3.15 & 3.13 & 3.40 & 3.88 & 3.21 & 2.90 & 1.49 & \textbf{1.00} \\
			& RMSE & 3.60 & 3.67 & 3.71 & 4.21 & 3.88 & 3.65 & 1.86 & \textbf{1.12} \\
			& MAE  & 2.09 & 2.07 & 2.39 & 2.57 & 2.13 & 1.93 & 0.99 & \textbf{0.67} \\
			Average
			 & MAPE & 4.16 & 4.47 & 4.25 & 5.01 & 4.96 & 4.07 & 2.34 & \textbf{1.75} \\
			 & RMSE & 4.27 & 4.35 & 4.30 & 5.04 & 4.96 & 4.22 & 2.53 & \textbf{1.72} \\
			 & MAE  & 2.76 & 2.96 & 2.85 & 3.32 & 3.29 & 2.70 & 1.55 & \textbf{1.16} \\
			\midrule
		\end{tabular}
	\end{adjustbox}
\end{table}

\begin{table}[H]
	\centering
	\caption{Percentage improvement of Proposed Method vs. Other Benchmarking Methods.}
	\label{table:Percentage-improvement}
	\scriptsize
	\begin{adjustbox}{max width=\linewidth}
		\begin{tabular}{
				>{\centering\arraybackslash}p{1.2cm}  
				>{\centering\arraybackslash}p{1.6cm}  
				>{\centering\arraybackslash}p{1.7cm}  
				>{\centering\arraybackslash}p{1.7cm}  
				>{\centering\arraybackslash}p{1.7cm}  
				>{\centering\arraybackslash}p{1.7cm}  
				>{\centering\arraybackslash}p{1.7cm}  
				>{\centering\arraybackslash}p{1.7cm}  
				>{\centering\arraybackslash}p{1.7cm}  
			}
			\toprule
			Month & Metrics & 
			Proposed Method vs SVR & 
			Proposed Method vs ANN & 
			Proposed Method vs RF & 
			Proposed Method vs CNN & 
			Proposed Method vs ResNet & 
			Proposed Method vs LSTM & 
			Proposed Method vs VMD-LSTM \\
			\midrule
			May
			& $P_{\text{MAPE}}$ & 0.66 & 0.67 & 0.66 & 0.73 & 0.72 & 0.66 & 0.48 \\
			& $P_{\text{RMSE}}$ & 0.67 & 0.66 & 0.66 & 0.72 & 0.70 & 0.65 & 0.45 \\
			& $P_{\text{MAE}}$  & 0.66 & 0.67 & 0.66 & 0.73 & 0.72 & 0.66 & 0.48 \\
			Jun
			& $P_{\text{MAPE}}$ & 0.55 & 0.57 & 0.58 & 0.63 & 0.65 & 0.56 & 0.14 \\
			& $P_{\text{RMSE}}$ & 0.57 & 0.57 & 0.58 & 0.63 & 0.65 & 0.57 & 0.31 \\
			& $P_{\text{MAE}}$  & 0.55 & 0.57 & 0.58 & 0.63 & 0.65 & 0.56 & 0.14 \\
			Jul
			& $P_{\text{MAPE}}$ & 0.48 & 0.53 & 0.49 & 0.55 & 0.60 & 0.50 & 0.15 \\
			& $P_{\text{RMSE}}$ & 0.52 & 0.55 & 0.52 & 0.58 & 0.62 & 0.53 & 0.22 \\
			& $P_{\text{MAE}}$  & 0.48 & 0.53 & 0.49 & 0.55 & 0.60 & 0.50 & 0.15 \\
			Aug
			& $P_{\text{MAPE}}$ & 0.53 & 0.68 & 0.42 & 0.58 & 0.45 & 0.33 & 0.07 \\
			& $P_{\text{RMSE}}$ & 0.44 & 0.53 & 0.48 & 0.59 & 0.49 & 0.42 & 0.08 \\
			& $P_{\text{MAE}}$  & 0.54 & 0.68 & 0.43 & 0.58 & 0.46 & 0.34 & 0.07 \\
			Sep
			& $P_{\text{MAPE}}$ & 0.68 & 0.68 & 0.71 & 0.74 & 0.69 & 0.66 & 0.33 \\
			& $P_{\text{RMSE}}$ & 0.69 & 0.69 & 0.70 & 0.73 & 0.71 & 0.69 & 0.40 \\
			& $P_{\text{MAE}}$  & 0.68 & 0.68 & 0.72 & 0.74 & 0.69 & 0.65 & 0.32 \\
			Average
			& $P_{\text{MAPE}}$ & 0.60 & 0.61 & 0.59 & 0.65 & 0.65 & 0.57 & 0.25 \\
			& $P_{\text{RMSE}}$ & 0.60 & 0.61 & 0.60 & 0.66 & 0.65 & 0.59 & 0.32 \\
			& $P_{\text{MAE}}$  & 0.58 & 0.61 & 0.59 & 0.65 & 0.65 & 0.57 & 0.25 \\
			\bottomrule
		\end{tabular}
	\end{adjustbox}
\end{table}

\section{Discussion}
\label{sec:Discussion}

The comparative experiments confirm that the proposed Bayesian-optimized MVMD-LSTM achieves superior forecasting accuracy compared with other benchmarking methods. This improvement can be attributed to three main factors. First, LSTM is better suited to capture temporal dependencies than shallow learning models. Second, decomposition-based hybrids effectively mitigate nonlinear and nonstationarity, thereby improving prediction performance. Third, the proposed method uses multivariate decomposition to capture cross-source dependencies among wind, solar and wave power, while Bayesian optimization adaptively determines MVMD parameters, leading to more accurate feature extraction and enhanced generalization.

The results of the decomposition also revealed different characteristics between the three energy sources. Wind power is dominated by high-frequency oscillations caused by turbulence and rapidly changing conditions, which explains its random and unpredictable nature. Solar power exhibits a pronounced diurnal periodicity, reflecting the daily irradiance cycle influenced by sunlight and cloud cover. In contrast, wave power is characterized by smooth, low-frequency changes, reflecting the inertia and stability of marine systems. In addition to these individual characteristics, a coupling relationship emerges: wind and solar outputs complement each other through their contrasting variability patterns, while wave power, with its delayed and smooth characteristics , acts as a stabilizing buffer. These interactions show that the advantages of joint modeling are better than each energy source alone.

Finally, the agreement between predicted and observed curves demonstrates the effectiveness of combining MVMD with LSTM in capturing both diurnal patterns and ultra-short-term fluctuations. Although small deviations occur during abrupt transitions, the model remains robust in forecasting renewable energy outputs.

\section{Conclusion}
\label{sec:conclusion}

This study proposed a Bayesian-optimized MVMD-LSTM framework for ultra-short-term forecasting of integrated wind–solar–wave power generation. By combining MVMD, Bayesian optimization and LSTM, the method effectively captures cross-source dependencies and adaptively determines decomposition parameters, thereby overcoming limitations of conventional based on decomposition forecasting approaches. Experiments on hybrid renewable energy data demonstrate that the proposed method consistently outperforms both single forecasting models and existing hybrid baselines in terms of MAPE, RMSE and MAE. These results confirm its high prediction accuracy, robustness and practical relevance, provide essential information for decision-makers to optimize load tracking and device control. Future work will extend the framework by join meteorological factors, validating on different place datasets and exploring real-time deployment in operational power systems. These efforts are expected to further improve predictive performance and enhance the applicability of the proposed approach in practical smart grid operation scenarios.

\bibliographystyle{elsarticle-num} 
\bibliography{reference}

\begin{thebibliography}{10}
\expandafter\ifx\csname url\endcsname\relax
  \def\url#1{\texttt{#1}}\fi
\expandafter\ifx\csname urlprefix\endcsname\relax\def\urlprefix{URL }\fi
\expandafter\ifx\csname href\endcsname\relax
  \def\href#1#2{#2} \def\path#1{#1}\fi

\bibitem{tian2022global}
J.~Tian, L.~Yu, R.~Xue, S.~Zhuang, Y.~Shan, Global low-carbon energy transition
  in the post-covid-19 era, Applied energy 307 (2022) 118205.

\bibitem{hassan2024renewable}
Q.~Hassan, P.~Viktor, T.~J. Al-Musawi, B.~M. Ali, S.~Algburi, H.~M. Alzoubi,
  A.~K. Al-Jiboory, A.~Z. Sameen, H.~M. Salman, M.~Jaszczur, The renewable
  energy role in the global energy transformations, Renewable Energy Focus 48
  (2024) 100545.

\bibitem{widen2015variability}
J.~Wid{\'e}n, N.~Carpman, V.~Castellucci, D.~Lingfors, J.~Olauson, F.~Remouit,
  M.~Bergkvist, M.~Grabbe, R.~Waters, Variability assessment and forecasting of
  renewables: A review for solar, wind, wave and tidal resources, Renewable and
  Sustainable Energy Reviews 44 (2015) 356--375.

\bibitem{che2025impact}
E.~E. Che, K.~R. Abeng, C.~D. Iweh, G.~J. Tsekouras, A.~Fopah-Lele, The impact
  of integrating variable renewable energy sources into grid-connected power
  systems: Challenges, mitigation strategies, and prospects, Energies 18~(3)
  (2025) 1--31.

\bibitem{jurasz2020review}
J.~Jurasz, F.~Canales, A.~Kies, M.~Guezgouz, A.~Beluco, A review on the
  complementarity of renewable energy sources: Concept, metrics, application
  and future research directions, Solar energy 195 (2020) 703--724.

\bibitem{modu2023systematic}
B.~Modu, M.~P. Abdullah, A.~L. Bukar, M.~F. Hamza, A systematic review of
  hybrid renewable energy systems with hydrogen storage: Sizing, optimization,
  and energy management strategy, International Journal of Hydrogen Energy
  48~(97) (2023) 38354--38373.

\bibitem{khalid2024smart}
M.~Khalid, Smart grids and renewable energy systems: Perspectives and grid
  integration challenges, Energy Strategy Reviews 51 (2024) 101299.

\bibitem{wei2024optimization}
D.~Wei, Z.~Zhang, W.~Zhang, Y.~Yang, Z.~Yang, Optimization of multi-energy
  complementary power generation system configuration based on particle swarm
  optimization, Energy Reports 12 (2024) 2257--2269.

\bibitem{hasan2025state}
M.~Hasan, Z.~Mifta, S.~J. Papiya, P.~Roy, P.~Dey, N.~A. Salsabil, N.-U.-R.
  Chowdhury, O.~Farrok, A state-of-the-art comparative review of load
  forecasting methods: Characteristics, perspectives, and applications, Energy
  Conversion and Management: X (2025) 100922.

\bibitem{wang2024economics}
W.~Wang, Y.~Guo, D.~Yang, Z.~Zhang, J.~Kleissl, D.~van~der Meer, G.~Yang,
  T.~Hong, B.~Liu, N.~Huang, et~al., Economics of physics-based solar
  forecasting in power system day-ahead scheduling, Renewable and Sustainable
  Energy Reviews 199 (2024) 114448.

\bibitem{ahmed2019review}
A.~Ahmed, M.~Khalid, A review on the selected applications of forecasting
  models in renewable power systems, Renewable and Sustainable Energy Reviews
  100 (2019) 9--21.

\bibitem{wang2024multi}
G.~Wang, Z.~Zhang, J.~Lin, Multi-energy complementary power systems based on
  solar energy: A review, Renewable and Sustainable Energy Reviews 199 (2024)
  114464.

\bibitem{hong2020energy}
T.~Hong, P.~Pinson, Y.~Wang, R.~Weron, D.~Yang, H.~Zareipour, Energy
  forecasting: A review and outlook, IEEE Open Access Journal of Power and
  Energy 7 (2020) 376--388.

\bibitem{hong2019ultra}
D.~Hong, T.~Ji, M.~Li, Q.~Wu, Ultra-short-term forecast of wind speed and wind
  power based on morphological high frequency filter and double similarity
  search algorithm, International Journal of Electrical Power \& Energy Systems
  104 (2019) 868--879.

\bibitem{jung2014current}
J.~Jung, R.~P. Broadwater, Current status and future advances for wind speed
  and power forecasting, Renewable and Sustainable Energy Reviews 31 (2014)
  762--777.

\bibitem{anvari2016short}
M.~Anvari, G.~Lohmann, M.~W{\"a}chter, P.~Milan, E.~Lorenz, D.~Heinemann,
  M.~R.~R. Tabar, J.~Peinke, Short term fluctuations of wind and solar power
  systems, New Journal of Physics 18~(6) (2016) 063027.

\bibitem{yu2022ultra}
G.~Z. Yu, L.~Lu, B.~Tang, S.~Y. Wang, C.~Chung, Ultra-short-term wind power
  subsection forecasting method based on extreme weather, IEEE Transactions on
  Power Systems 38~(6) (2022) 5045--5056.

\bibitem{lafuente2025state}
M.~Lafuente-Cacho, O.~Izquierdo-Monge, P.~Pe{\~n}a-Carro,
  {\'A}.~Hern{\'a}ndez-Jim{\'e}nez, L.~H. Callejo, A.~M.~P. Losada,
  {\'A}.~L.~Z. Lamadrid, State of the art for solar and wind energy-forecasting
  methods for sustainable grid integration, Current Sustainable/Renewable
  Energy Reports 12~(1) (2025) 1--12.

\bibitem{yu2024ultra}
G.~Yu, L.~Shen, Q.~Dong, G.~Cui, S.~Wang, D.~Xin, X.~Chen, W.~Lu,
  Ultra-short-term wind power forecasting techniques: comparative analysis and
  future trends, Frontiers in Energy Research 11 (2024) 1345004.

\bibitem{hassan2023review}
Q.~Hassan, S.~Algburi, A.~Z. Sameen, H.~M. Salman, M.~Jaszczur, A review of
  hybrid renewable energy systems: Solar and wind-powered solutions:
  Challenges, opportunities, and policy implications, Results in engineering 20
  (2023) 101621.

\bibitem{guo2021review}
B.~Guo, J.~V. Ringwood, A review of wave energy technology from a research and
  commercial perspective, IET Renewable Power Generation 15~(14) (2021)
  3065--3090.

\bibitem{alsamamra2024performance}
H.~R. Alsamamra, S.~Salah, J.~H. Shoqeir, Performance analysis of arima model
  for wind speed forecasting in jerusalem, palestine, Energy Exploration \&
  Exploitation 42~(5) (2024) 1727--1746.

\bibitem{chodakowska2023arima}
E.~Chodakowska, J.~Nazarko, {\L}.~Nazarko, H.~S. Rabayah, R.~M. Abendeh,
  R.~Alawneh, Arima models in solar radiation forecasting in different
  geographic locations, Energies 16~(13) (2023) 5029.

\bibitem{reikard2009forecasting}
G.~Reikard, Forecasting ocean wave energy: Tests of time-series models, Ocean
  Engineering 36~(5) (2009) 348--356.

\bibitem{atecs2023estimation}
K.~T. Ate{\c{s}}, Estimation of short-term power of wind turbines using
  artificial neural network (ann) and swarm intelligence, Sustainability
  15~(18) (2023) 13572.

\bibitem{ding2011ann}
M.~Ding, L.~Wang, R.~Bi, An ann-based approach for forecasting the power output
  of photovoltaic system, Procedia Environmental Sciences 11 (2011) 1308--1315.

\bibitem{hadadpour2014wave}
S.~Hadadpour, A.~Etemad-Shahidi, B.~Kamranzad, Wave energy forecasting using
  artificial neural networks in the caspian sea, in: Proceedings of the
  institution of civil engineers-maritime engineering, Vol. 167, Thomas Telford
  Ltd, 2014, pp. 42--52.

\bibitem{yuan2022wind}
D.-D. Yuan, M.~Li, H.-Y. Li, C.-J. Lin, B.-X. Ji, Wind power prediction method:
  Support vector regression optimized by improved jellyfish search algorithm,
  Energies 15~(17) (2022) 6404.

\bibitem{das2017svr}
U.~K. Das, K.~S. Tey, M.~Seyedmahmoudian, M.~Y. Idna~Idris, S.~Mekhilef,
  B.~Horan, A.~Stojcevski, Svr-based model to forecast pv power generation
  under different weather conditions, Energies 10~(7) (2017) 876.

\bibitem{zhou2016wind}
Z.~Zhou, X.~Li, H.~Wu, Wind power prediction based on random forests, in: 2016
  4th International Conference on Electrical \& Electronics Engineering and
  Computer Science (ICEEECS 2016), Atlantis Press, 2016, pp. 352--356.

\bibitem{guan2024study}
L.~Guan, L.~Zou, Study on solar power prediction model by random forest method
  based on a numerical weather prediction model, Authorea Preprints (2024).

\bibitem{qu2021day}
J.~Qu, Z.~Qian, Y.~Pei, Day-ahead hourly photovoltaic power forecasting using
  attention-based cnn-lstm neural network embedded with multiple relevant and
  target variables prediction pattern, Energy 232 (2021) 120996.

\bibitem{mirza2023hybrid}
A.~F. Mirza, M.~Mansoor, M.~Usman, Q.~Ling, Hybrid inception-embedded deep
  neural network resnet for short and medium-term pv-wind forecasting, Energy
  Conversion and Management 294 (2023) 117574.

\bibitem{zhang1998forecasting}
G.~Zhang, B.~E. Patuwo, M.~Y. Hu, Forecasting with artificial neural networks::
  The state of the art, International journal of forecasting 14~(1) (1998)
  35--62.

\bibitem{zhang2019wind}
J.~Zhang, X.~Jiang, X.~Chen, X.~Li, D.~Guo, L.~Cui, Wind power generation
  prediction based on lstm, in: Proceedings of the 2019 4th international
  conference on Mathematics and artificial intelligence, 2019, pp. 85--89.

\bibitem{konstantinou2021solar}
M.~Konstantinou, S.~Peratikou, A.~G. Charalambides, Solar photovoltaic
  forecasting of power output using lstm networks, Atmosphere 12~(1) (2021)
  124.

\bibitem{minuzzi2023deep}
F.~C. Minuzzi, L.~Farina, A deep learning approach to predict significant wave
  height using long short-term memory, Ocean Modelling 181 (2023) 102151.

\bibitem{chen2023regional}
Y.~Chen, J.-W. Xiao, Y.-W. Wang, Y.~Li, Regional wind-photovoltaic combined
  power generation forecasting based on a novel multi-task learning framework
  and tpa-lstm, Energy Conversion and Management 297 (2023) 117715.

\bibitem{sinsel2020challenges}
S.~R. Sinsel, R.~L. Riemke, V.~H. Hoffmann, Challenges and solution
  technologies for the integration of variable renewable energy sources—a
  review, renewable energy 145 (2020) 2271--2285.

\bibitem{qian2019review}
Z.~Qian, Y.~Pei, H.~Zareipour, N.~Chen, A review and discussion of
  decomposition-based hybrid models for wind energy forecasting applications,
  Applied energy 235 (2019) 939--953.

\bibitem{dragomiretskiy2013variational}
K.~Dragomiretskiy, D.~Zosso, Variational mode decomposition, IEEE transactions
  on signal processing 62~(3) (2013) 531--544.

\bibitem{wu2022completed}
Y.-K. Wu, C.-L. Huang, Q.-T. Phan, Y.-Y. Li, Completed review of various solar
  power forecasting techniques considering different viewpoints, Energies
  15~(9) (2022) 3320.

\bibitem{gu2022review}
C.~Gu, H.~Li, Review on deep learning research and applications in wind and
  wave energy, Energies 15~(4) (2022) 1510.

\bibitem{ghanbari2025short}
E.~Ghanbari, A.~Avar, Short-term wind power forecasting using the hybrid model
  of multivariate variational mode decomposition (mvmd) and long short-term
  memory (lstm) neural networks, Electrical Engineering 107~(3) (2025)
  2903--2933.

\bibitem{ur2019multivariate}
N.~ur~Rehman, H.~Aftab, Multivariate variational mode decomposition, IEEE
  Transactions on signal processing 67~(23) (2019) 6039--6052.

\bibitem{eriksen2022data}
T.~Eriksen, et~al., Data-driven signal decomposition approaches: A comparative
  analysis, arXiv preprint arXiv:2208.10874 (2022).

\bibitem{zulfiqar2023hybrid}
M.~Zulfiqar, M.~Kamran, M.~Rasheed, T.~Alquthami, A.~Milyani, A hybrid
  framework for short term load forecasting with a navel feature engineering
  and adaptive grasshopper optimization in smart grid, Applied Energy 338
  (2023) 120829.

\bibitem{tang2023novel}
Q.~Tang, Y.~Jiang, J.~Xin, G.~Liao, J.~Zhou, X.~Yang, A novel method for the
  recovery of continuous missing data using multivariate variational mode
  decomposition and fully convolutional networks, Measurement 220 (2023)
  113366.

\bibitem{shahriari2015taking}
B.~Shahriari, K.~Swersky, Z.~Wang, R.~P. Adams, N.~De~Freitas, Taking the human
  out of the loop: A review of bayesian optimization, Proceedings of the IEEE
  104~(1) (2015) 148--175.

\bibitem{frazier2018tutorial}
P.~I. Frazier, A tutorial on bayesian optimization, arXiv preprint
  arXiv:1807.02811 (2018).

\bibitem{snoek2012practical}
J.~Snoek, H.~Larochelle, R.~P. Adams, Practical bayesian optimization of
  machine learning algorithms, Advances in neural information processing
  systems 25 (2012).

\bibitem{wu2017bayesian}
J.~Wu, M.~Poloczek, A.~G. Wilson, P.~Frazier, Bayesian optimization with
  gradients, Advances in neural information processing systems 30 (2017).

\bibitem{wang2016bayesian}
Z.~Wang, F.~Hutter, M.~Zoghi, D.~Matheson, N.~De~Feitas, Bayesian optimization
  in a billion dimensions via random embeddings, Journal of Artificial
  Intelligence Research 55 (2016) 361--387.

\bibitem{habtemariam2023bayesian}
E.~T. Habtemariam, K.~Kekeba, M.~Mart{\'\i}nez-Ballesteros,
  F.~Mart{\'\i}nez-{\'A}lvarez, A bayesian optimization-based lstm model for
  wind power forecasting in the adama district, ethiopia, Energies 16~(5)
  (2023) 2317.

\bibitem{guangzhou2023offshore}
{Guangzhou Institute of Energy Conversion, Chinese Academy of Sciences},
  Dataset of offshore multi-energy complementary power generation system,
  \url{https://cstr.cn/16666.11.nbsdc.MbYZ7mLV} (2023).

\bibitem{akiba2019optuna}
T.~Akiba, S.~Sano, T.~Yanase, T.~Ohta, M.~Koyama, Optuna: A next-generation
  hyperparameter optimization framework, in: Proceedings of the 25th ACM SIGKDD
  international conference on knowledge discovery \& data mining, 2019, pp.
  2623--2631.

\bibitem{Dmocrito2024mvmd}
Dmocrito, Multivariate variational mode decomposition,
  \url{https://github.com/Dmocrito/mvmd} (2024).

\bibitem{hochreiter1997long}
S.~Hochreiter, J.~Schmidhuber, Long short-term memory, Neural computation 9~(8)
  (1997) 1735--1780.

\bibitem{kinga2015method}
D.~Kinga, J.~B. Adam, et~al., A method for stochastic optimization, in:
  International conference on learning representations (ICLR), Vol.~5,
  California;, 2015.

\bibitem{karijadi2023wind}
I.~Karijadi, S.-Y. Chou, A.~Dewabharata, Wind power forecasting based on hybrid
  ceemdan-ewt deep learning method, Renewable Energy 218 (2023) 119357.

\bibitem{liu2020new}
H.~Liu, C.~Yu, H.~Wu, Z.~Duan, G.~Yan, A new hybrid ensemble deep reinforcement
  learning model for wind speed short term forecasting, Energy 202 (2020)
  117794.

\end{thebibliography}

\end{document}